\documentclass[runningheads]{llncs}
\usepackage{makeidx}
\usepackage{graphicx}
\usepackage{amsmath,amssymb} 
\usepackage{color}
\usepackage{wrapfig}
\usepackage{sidecap}
\usepackage{subcaption}
\usepackage{multirow}
\usepackage{makecell}
\usepackage{bbold}

\usepackage{hyperref}
\newcommand{\myparagraph}[1]{\noindent{\bf #1}}

\begin{document}
	\pagestyle{headings}
	\mainmatter

	\title{Long-Tailed Recognition Using Class-Balanced Experts}

 \author{Saurabh Sharma\inst{1},
            Ning Yu\inst{1,2},
            Mario Fritz\inst{3}, Bernt Schiele\inst{1}}
            \authorrunning{S. Sharma et al.}
            \institute{Max Planck Institute for Informatics, Saarland Informatics Campus \and University of Maryland, College Park \and CISPA Helmholtz Center for Information Security, Saarland Informatics Campus
            \email{\{ssharma,ningyu,schiele\}@mpi-inf.mpg.de}$\;$ \email{fritz@cispa.saarland}}
	\maketitle
    \vspace{-2em}

\begin{abstract}
Deep learning enables impressive performance in image recognition using large-scale artificially-balanced datasets. However, real-world datasets exhibit highly class-imbalanced distributions, yielding two main challenges: relative imbalance amongst the classes and data scarcity for mediumshot or fewshot classes. In this work, we address the problem of long-tailed recognition wherein the training set is highly imbalanced and the test set is kept balanced. Differently from existing paradigms relying on data-resampling, cost-sensitive learning, online hard example mining, loss objective reshaping, and/or memory-based modeling, we propose an ensemble of class-balanced experts that combines the strength of diverse classifiers.
Our ensemble of class-balanced experts reaches results close to state-of-the-art and an extended ensemble establishes a new state-of-the-art on two benchmarks for long-tailed recognition. We conduct extensive experiments to analyse the performance of the ensembles, and discover that in modern large-scale datasets, relative imbalance is a harder problem than data scarcity. The training and evaluation code is available at \url{https://github.com/ssfootball04/class-balanced-experts}.
\end{abstract}

\section{Introduction}
In the past decades, deep learning has boosted success in image recognition to a new level~\cite{he2016deep}. The availability of large-scale datasets with thousands of images in each class~\cite{deng2009imagenet,zhou2017places} has been a major factor in this revolution. However, these datasets are manually curated and artificially balanced, as opposed to real-world datasets that exhibit a highly skewed and class-imbalanced distribution in a long-tailed shape: a few common classes and many more rare classes. To address this practical challenge, in this work, we focus on the problem of long-tailed recognition, wherein datasets exhibit a natural power-law distribution~\cite{van2017devil}, allowing us to assess model performance on four folds: \emph{Manyshot} classes ($\geq100$ samples), \emph{Mediumshot} classes ($20\sim100$ samples), \emph{Fewshot} classes ($<20$ samples), and \emph{All} classes. Training data follows a highly class-imbalanced distribution, and testing data is balanced so that equally good performance over all classes is crucial~\cite{liu2019large}.

The two main challenges for a long-tailed classification model are \emph{relative imbalance} amongst the classes, and \emph{data scarcity} or unobservable data modes~\cite{he2009learning}. Existing techniques for imbalanced classification have focused on data re-sampling ~\cite{estabrooks2004multiple,he2009learning} and cost-sensitive learning~\cite{lin2017focal,cui2019class} to re-weigh the loss objective or counter \emph{relative imbalance}, while techniques for fewshot learning have employed data augmentation~\cite{felix2018multi,wang2018low,xian2018feature,xian2019f}, classifier weight prediction~\cite{gidaris2018dynamic,qiao2018few,qi2018low}, or prototype-based non-parametric methods~\cite{snell2017prototypical,vinyals2016matching,liu2019large} to address \emph{data scarcity}. 

Unlike the aforementioned paradigms, we instead revisit the classic approach of ensemble of experts \cite{jacobs1991adaptive,yuksel2012twenty,hinton2015distilling} and adapt it to long-tailed recognition. We first decompose the imbalanced classification problem into balanced classification problems by splitting the long-tailed training classes into balanced subsets. Then we train an expert on each balanced subset, so-called \emph{Manyshot}, \emph{Mediumshot}, or \emph{Fewshot} data, with out-of-distribution detection for samples outside an expert's class-balanced subset. This explicitly tackles the issue of \emph{relative imbalance}, and prevents competition between \emph{Manyshot} and \emph{Fewshot} classes during training. 

Further, to use all available data for learning feature representations and to \emph{transfer knowledge} from \emph{Manyshot} to \emph{Mediumshot} and \emph{Fewshot} classes, we initialise the feature extractor of each expert using a \emph{Baseline} model trained on the entire dataset. This simple and effective approach reaches close to state-of-the-art results without involving more complex models or sophisticated loss objectives. Moreover, the decomposition into class-balanced subsets allows us to analyse the upper bound on performance in each data regime. Specifically, our experiments with an \emph{Oracle} upper bound allow us to bring \emph{Fewshot} and \emph{Mediumshot} accuracy on par with \emph{Manyshot} accuracy, revealing that in modern large-scale datasets the data scarcity for \emph{Mediumshot} and \emph{Fewshot} classes can be effectively handled using knowledge transfer from \emph{Manyshot} classes. Therefore, relative imbalance is a more severe problem.

\begin{figure}[!t]
        \centering
        \includegraphics[width=\linewidth]{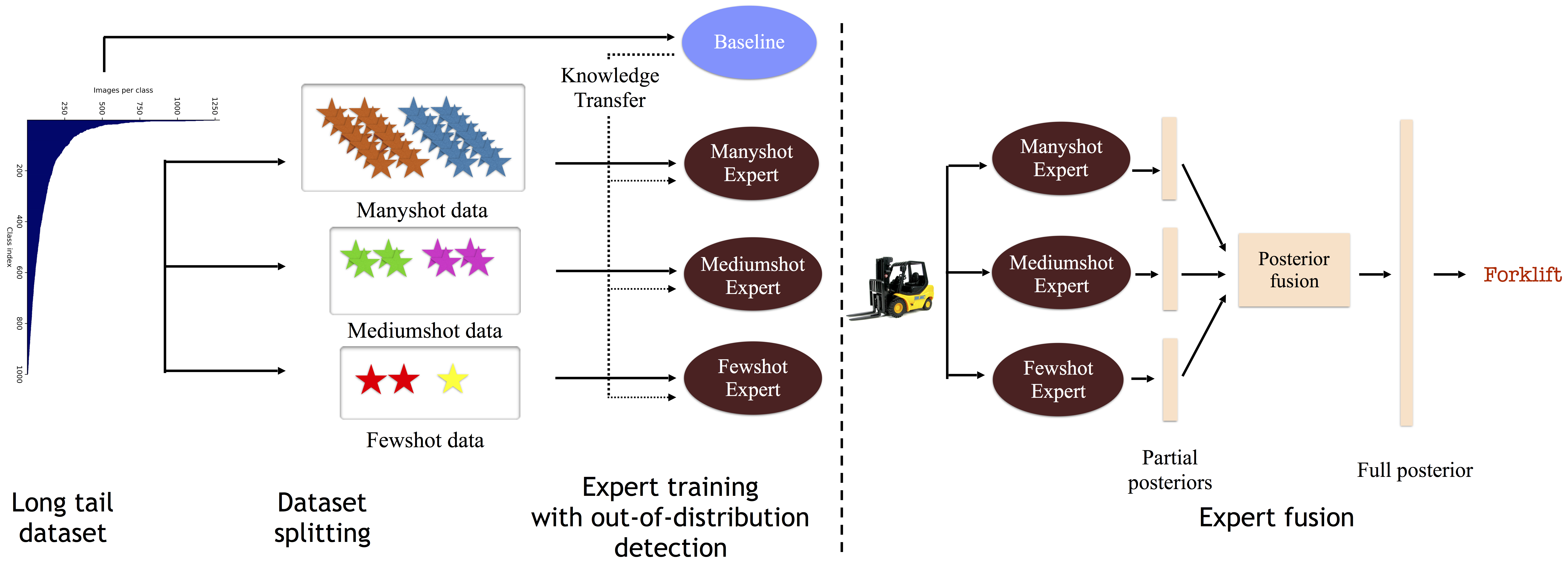}
        \caption{Our pipeline for long-tailed recognition: an ensemble of experts trained on class-balanced subsets of \emph{Manyshot}, \emph{Mediumshot}, and \emph{Fewshot} data. We \emph{transfer knowledge} from \emph{Manyshot} to \emph{Mediumshot} and \emph{Fewshot} classes by initialising experts with a \emph{Baseline} model trained on all the data. Expert models classify samples outside their subset as out-of-distribution and output partial posteriors that are fused into a full posterior to obtain the final prediction.}
        \label{fig:teaser}
\end{figure}

We also leverage the flexibility and modularity of the ensemble framework to create larger and more diverse ensembles using existing solutions for long-tailed recognition. In particular, we involve the following methods in the solution space: (1) a \emph{Baseline} model without any bells or whistles; (2) feature learning followed by classifier finetuning with uniform class sampling~\cite{sun2019meta,xian2019f}; (3) data augmentation using feature generation networks~\cite{felix2018multi,wang2018low,xian2019f}; and (4) knowledge transfer through prototype-based memory representation~\cite{snell2017prototypical,liu2019large}. The extended ensemble consisting of all these models outperforms the current state-of-the-art on two benchmark datasets by a significant margin.

Our \textbf{contributions} in this work can be summarised as follows:

(1) We propose an effective and modular ensemble of experts framework for long-tailed recognition that decomposes the imbalanced classification problem into multiple balanced classification problems. Our framework utilises all available data for learning feature representations and transfers this knowledge from \emph{Manyshot} to \emph{Mediumshot} and \emph{Fewshot} classes. The results of our ensemble of class-balanced experts are close to the state-of-the-art performance on two long-tailed benchmark datasets, ImageNet-LT and Places-LT~\cite{liu2019large}.

(2) We enrich our ensemble with a diverse set of existing solutions for long-tailed recognition, namely data re-sampling, data augmentation using synthesised features, and prototype-based classification, and establish a new state-of-the-art for long-tailed recognition. 

(3) We analyse the upper bound performance of our approach in the following manner: we assume Oracle access to the experts containing the ground truth classes of the test samples in their class-balanced subsets. We discover that \emph{data scarcity} for rare classes is not a severe issue in modern large-scale datasets. Rather, \emph{relative imbalance} is the main bottleneck.

\section{Related work}
\myparagraph{Imbalanced classification and long-tailed recognition.} There is a long history of research in imbalanced classification~\cite{he2009learning,bengio2015battle,van2017devil}, in binary and more generally multi-class classification problems. Classic problems that naturally encounter class imbalance are face attribute detection~\cite{oh2016deep,huang2019deep}, object detection~\cite{zhu2014capturing,lin2017focal}, and image defect detection~\cite{yu2018learning}. Prior work on image classification~\cite{wang2016learning,wang2017learning} deals with long-tailed datasets, but only recently a benchmark for the problem on the ImageNet and Places dataset was proposed by~\cite{liu2019large}. They also propose splits for open-world classification, but in this work we only consider long-tailed recognition and we report the performance of our methods on the proposed ImageNet-LT and Places-LT. We summarise below the existing solutions for imbalanced classification and long-tailed recognition.

\myparagraph{Data re-sampling heuristics and cost-sensitive learning.} These are classic ways to tackle long-tailed recognition. A more balanced data distribution is achieved by randomly over-sampling fewshot classes or randomly under-sampling of manyshot classes~\cite{estabrooks2004multiple,he2009learning}. However, over-sampling suffers from overfitting on fewshot classes while under-sampling cannot take full benefit of available data for generalization on manyshot classes. Other work has focused on hard example mining~\cite{dong2017class} or cost-sensitive learning~\cite{lin2017focal,cui2019class} reasoned from class frequencies. Instead, to augment our ensemble of class-balanced experts, we use a uniform class sampling procedure in mini-batch training for finetuning the classifier after a representation learning phase, which has the advantage that all data is used to learn representations while decision boundary learning takes class imbalance into account. This has also been employed before in related zero-shot learning~\cite{xian2019f} and fewshot learning~\cite{sun2019meta} work.

\myparagraph{Synthetic data augmentation.} This is a classic technique that synthesises features for minority classes based on feature space similarities~\cite{chawla2002smote,han2005borderline}. More recently, generative models have been employed in zero-shot~\cite{felix2018multi,xian2018feature,xian2019f} and fewshot learning~\cite{wang2018low} literature to automatically generate images or feature embeddings for data-starved classes. In this work, we use the f-VAEGAN-D2 model from~\cite{xian2019f} that generates feature embeddings conditioned on available class embeddings using a VAE-GAN model, and integrate it into our ensemble of experts framework.

\myparagraph{Prototype-based models and knowledge transfer.} Prototype-based networks~\cite{snell2017prototypical,vinyals2016matching} maintain a memory module for all the classes such that each class is equally represented regardless of sample frequency. In particular, Liu~et~al.~\cite{liu2019large} learn prototype-based features on-the-fly to effectively transfer knowledge from manyshot classes to fewshot classes. We integrate their model into our ensemble due to its ability to perform consistently well across the entire class spectrum. Transfer learning \cite{oquab2014learning} addresses data imbalance by transferring abundant features of manyshot classes to those of fewshot classes. Recent work includes transferring the intra-class variance~\cite{yin2019feature} and transferring semantic deep features~\cite{zhong2019unequal,liu2019large}. We instead transfer knowledge across the dataset by initialising our expert models with a baseline model pre-trained on the entire dataset.

\myparagraph{Ensemble learning.} Ensemble methods are a well-studied topic in machine learning literature. In particular, a variety of ensemble-based methods using boosting~\cite{guo2004learning,wang2010boosting}, bagging~\cite{galar2011review,khoshgoftaar2010comparing}, stacking~\cite{wang2009diversity}, and evolutionary selection of classifiers~\cite{krawczyk2014cost} have been employed for imbalanced datasets. However, they all consider ensembles with the same kind of model and task. Our approach is related to the work of Hinton~et~al.~\cite{hinton2015distilling} who train an ensemble of experts over disjoint semantically-close subsets of classes, thereby each expert deals with a different classification task. We instead train our experts on subsets of classes that are intrinsically balanced to counter relative imbalance and prevent competition between manyshot and fewshot classes during training. Moreover, we integrate a diverse set of models for long-tailed recognition into our ensemble of experts.

\myparagraph{Out-of-distribution detection and confidence calibration.} Modern neural networks can function both as classification models and detectors for out-of-distribution examples \cite{hendrycks17baseline}. Recent works focus on adding small perturbations in input space and applying temperature scaling \cite{liang2018enhancing}, and adding loss terms to push out-of-distribution examples towards uniform confidence \cite{hendrycks2019oe}. Related work on confidence calibration tries to fix overconfident predictions on in-distribution data using temperature scaling \cite{guo2017calibration}. We instead focus on learning an ensemble of class-balanced experts for long-tailed recognition, where the problem of out-of-distribution detection arises when dealing with samples from outside an expert's subset, and jointly calibrate experts' confidences to fuse their posteriors.

\section{Method}
We propose an ensemble of experts for solving the problem of long-tailed recognition. We split the long-tailed dataset into (approximately) class-balanced subsets, and a separate classification model, or expert, is trained for each subset. Expert models identify samples belonging to classes outside their subset as out-of-distribution; therefore we train them to produce low confidence predictions on these samples. During inference, each classification model yields a partial posterior distribution for test samples, the ensemble of which is fused to form a complete posterior distribution. Our entire pipeline is depicted in Fig.~\ref{fig:teaser}. The modularity of our framework allows us to explictly address the problem of \emph{relative imbalance}, and moreover analyse the upper bounds for performance in each data regime using Oracle access to experts containing ground truth classes of test samples in their class-balanced subsets.

\subsection{Long-tailed recognition using class-balanced experts}
\label{Long-tailed recognition using class-balanced experts}
The task of long-tailed visual recognition is as follows: given class-imbalanced training set $\mathcal{D}_\textit{Train} = \{(x_i,y_i)\}_{i=1}^n$ and class-balanced validation set $\mathcal{D}_\textit{Val}$ and class-balanced test set $\mathcal{D}_\textit{Test}$, the objective is to maximise test accuracy on four folds,
\emph{Manyshot} classes ($\geq100$ samples), \emph{Mediumshot} classes ($20\sim100$ samples), \emph{Fewshot} classes ($<20$ samples), and \emph{All} classes. This is a hard problem, since any high performing model must deal with the two problems of relative imbalance and data scarcity. 

\begin{wrapfigure}{l}{0.4\linewidth}
        \vspace{-2em}
        \includegraphics[width=\linewidth]{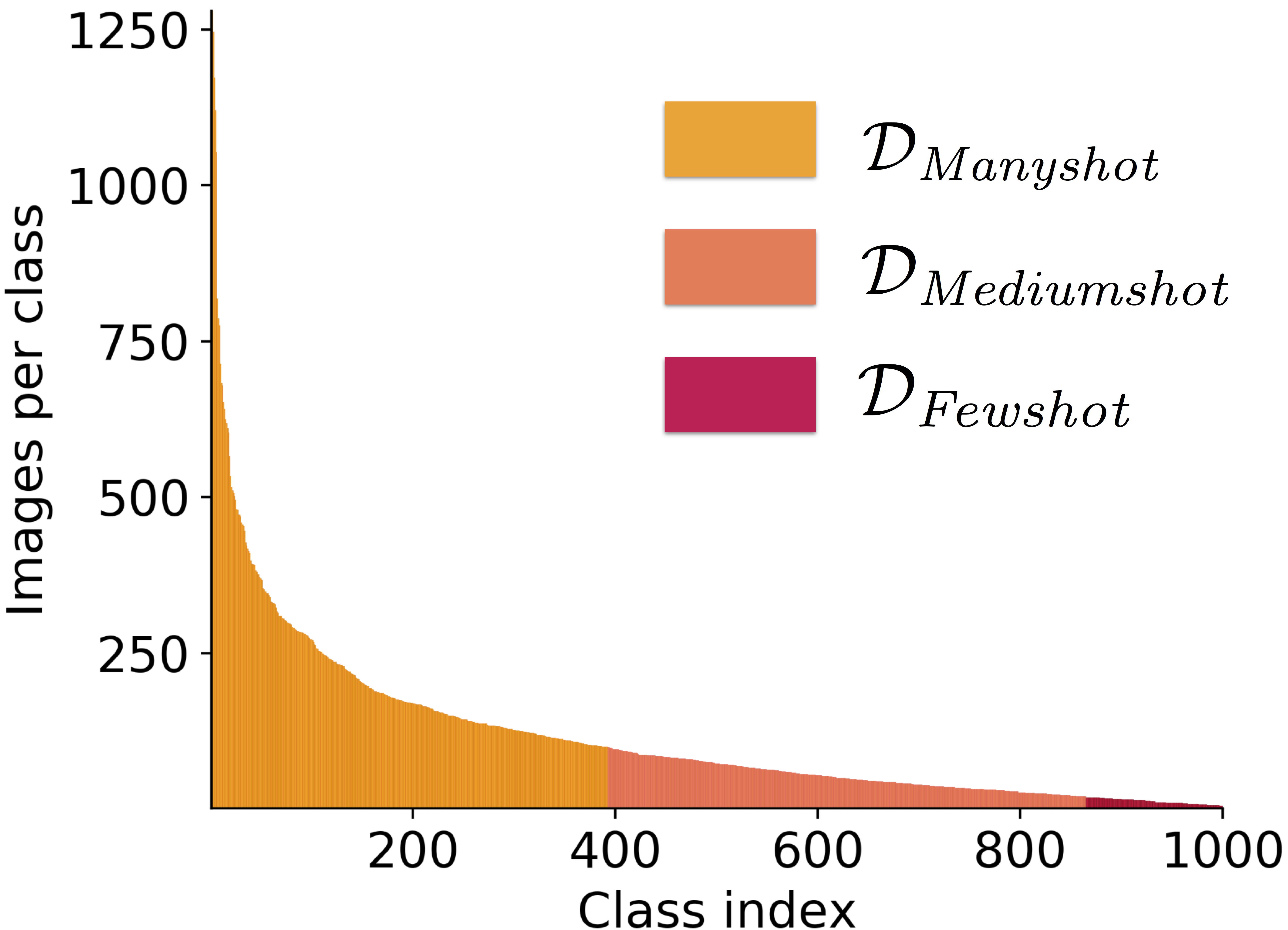}
        \vspace{-1em}
        \caption{Dataset splitting: We decompose ImageNet-LT into (relatively) class-balanced \emph{Manyshot}, \emph{Mediumshot}, and \emph{Fewshot} data subsets.}
        \vspace{-2em}
        \label{fig:dataset_splitting}
\end{wrapfigure}

Relative imbalance leads to biased classification boundaries wherein accuracy on fewshot samples is compromised in favor of manyshot samples that dominate the training objective. Data scarcity leads to representations that do not model unobserved data modes and is more severe. To tackle both these issues, we sort the class-imbalanced training set $\mathcal{D}_\textit{Train}$ according to class frequencies and partition it into contiguous class-balanced subsets $\mathcal{D}_\textit{Manyshot}$, $\mathcal{D}_\textit{Mediumshot}$ and $\mathcal{D}_\textit{Fewshot}$. This is visualised in Fig.~\ref{fig:dataset_splitting}. \\ For each subset, we train separate classification models or experts, that are initialised using a model pre-trained on the entire dataset. 
 Consequently we obtain the expert models $\mathcal{E}_\textit{Manyshot}$, $\mathcal{E}_\textit{Mediumshot}$ and $\mathcal{E}_\textit{Fewshot}$ corresponding to each class-balanced subset. The feature extractor part of each expert model $\mathcal{E}_{\_}$ is initialised using the \emph{Baseline} model pre-trained on the entire training set $\mathcal{D}_\textit{Train}$. This enables knowledge transfer from \emph{Manyshot} to \emph{Mediumshot} and \emph{Fewshot} classes. In this work, the expert models $\mathcal{E}_{\_}$ and the \emph{Baseline} model are deep fully convolutional neural networks with softmax classifiers.

\subsection{Out-of-distribution detection for experts}
\label{Out-of-distribution detection for experts}
The expert models identify samples from classes outside their class-balanced subset as out-of-distribution or OOD for short, therefore we train them using an out-of-distribution detection strategy. Observe that this is a hard problem, since here OOD examples come from within the same distribution albeit from extra classes within the dataset, as opposed to standard out-of-distribution detection wherein OOD samples come from an entirely different dataset.
\subsubsection{Training with reject class.}
We add a reject class to the softmax classifier of each expert. For instance, $\mathcal{E}_\textit{Manyshot}$ treats samples from $\mathcal{D}_\textit{Mediumshot} \cup \mathcal{D}_\textit{Fewshot}$ as a single reject class. This introduces imbalance since the reject class has far more samples than any other class, therefore we undersample reject class samples appropriately during training. We correct for the statistical bias by incrementing its logit score by the log of the undersampling ratio. We note that samples in the reject class have very high variance and are therefore hard to fit.  

\subsection{Fusing expert posteriors}
We consider various baseline strategies and propose a novel joint calibration module to fuse expert posteriors $\mathcal{E}_{\_}(x)$ into a complete posterior distribution. The final prediction and confidence scores are taken from this posterior, denoted as $q(x)$, using the argmax operation.

\label{Fusing expert predictions}
\subsubsection{KL-divergence minimisation.} We find the full posterior distribution for each sample, by minimising its KL-divergence with all the partial posterior distributions predicted by the experts \cite{hinton2015distilling}, that is,  
\begin{center}
    $\displaystyle \min_{q(x)} \sum_{\mathcal{E}_{\_}} KL(\mathcal{E}_{\_}(x)||q(x))$
\end{center}
where $q(x)$ is parameterised using logits $z$ and a softmax function as $q(x) = softmax(z)$. Note that probabilities corresponding to out-of-distribution classes for the expert $\mathcal{E}_{\_}$ are summed up into one probability score in $q(x)$ to align the two distributions. 

\subsubsection{Soft-voting.} We find the full posterior by summing up the partial posteriors directly and normalising the sum to 1, 
\begin{center}
        $q(x) = \frac{\displaystyle\sum_{\mathcal{E}_{\_}} g(\mathcal{E}_{\_}(x))}{\displaystyle\sum_{\mathcal{E}_{\_}} \mathbb{1}}$
\end{center}
Here $g(.)$ is a function that converts an expert's partial posterior into a full posterior. Since experts are trained with a reject class, $g(.)$ averages reject class probability score across out-of-distribution classes corresponding to expert $\mathcal{E}_{\_}$.

\subsubsection{Expert selection.} We train a 3-way classifier on the validation set, taking the partial posterior vectors $\mathcal{E}_{\_}(x)$ of each expert $\mathcal{E}_{\_}$ as input, to predict for a sample $x$ the expert model $\mathcal{E}_{\_}$ that contain's the sample's ground truth class in its class-balanced subset. Thus, for instance, the classifier learns to predict that a manyshot sample lies in the class-balanced subset of the manyshot expert $\mathcal{E}_\textit{Manyshot}$. The full posterior $q(x)$ is then given by $g(\mathcal{E}_{\_}(x))$ for the predicted expert $\mathcal{E}_{\_}$, where $g(.)$ is defined similarly as before.    
\subsubsection{Model stacking.} We train a single layer linear softmax classifier to predict the full posterior q(x) from the partial posterior vectors $\mathcal{E}_{\_}(x)$ of each expert $\mathcal{E}_{\_}$. The vectors $\mathcal{E}_{\_}(x)$ are concatenated to form a feature embedding for the softmax classifier which is trained by optimising the cross entropy loss on the validation set. This is a standard way for ensemble fusion known as model stacking~\cite{wolpert1992stacked}. 
\subsubsection{Joint calibration.} We calibrate the partial posteriors $\mathcal{E}_{\_}(x)$ by learning scaling and shift parameters before adding up the posteriors similarly to soft-voting,        
\begin{center}
        $q(x) = \frac{\displaystyle\sum_{\mathcal{E}_{\_}} g(\sigma_{SM} (w_{\mathcal{E}_{\_}} \odot z_{\mathcal{E}_{\_}}(x) + b_{\mathcal{E}_{\_}}))}{\mathbb{Z}}$
\end{center}
where $\sigma_{SM}$ denotes the softmax operation, $w_{\mathcal{E}_{\_}}$ and $b_{\mathcal{E}_{\_}}$ are scale and shift parameters respectively,  $z_{\mathcal{E}_{\_}}(x)$ denotes the logit scores of expert $\mathcal{E}_{\_}$ for sample $x$, $\odot$ denotes elementwise multiplication of two vectors, $\mathbb{Z}$ is a normalisation factor, and $g(.)$ is defined as before. We learn scale and shift parameters by minimising the cross entropy loss on the validation set. This module effectively learns the right alignment for experts' partial posteriors before performing soft-voting.

\section{Experiments}

\myparagraph{Datasets.} We use the object-centric ImageNet-LT and scene-centric Places-LT datasets for long-tailed recognition, released by Liu~et~al.~\cite{liu2019large}. The training set statistics are depicted in Table~\ref{tab:dataset_statistics}. ImageNet-LT has an imbalanced training set with 115,846 images for 1,000 classes from ImageNet-1K~\cite{deng2009imagenet}.

\begin{wraptable}{l}{.55\linewidth}
\vspace{-2em}
\scriptsize
\caption{Statistics for training sets in ImageNet-LT and Places-LT.}
\begin{tabular}{ l | l | c c c c }
\Xhline{2\arrayrulewidth}
Datasets & Attributes & Many & Medium & Few & All \\
\Xhline{2\arrayrulewidth}
ImageNet-LT & Classes & 391 & 473 & 136 & 1,000 \\
 & Samples & 89,293 & 24,910 & 1,643 & 115,846 \\ \hline
Places-LT & Classes & 132 & 162 & 71 & 365 \\
 & Samples & 52,862 & 8,834 & 804 & 62,500 \\
\Xhline{2\arrayrulewidth}
\end{tabular}
\vspace{-2em}
\label{tab:dataset_statistics}
\end{wraptable}

The class frequencies follow a natural power-law distribution~\cite{van2017devil} with a maximum number of 1,280 images per class and a minimum number of 5 images per class. The validation and testing sets are balanced and contain 20 and 50 images per class respectively. Places-LT has an imbalanced training set with 62,500 images for 365 classes from Places-2~\cite{zhou2017places}. The class frequencies follow a natural power-law distribution~\cite{van2017devil} with a maximum number of 4,980 images per class and a minimum number of 5 images per class. The validation and testing sets are balanced and contain 20 and 100 images per class respectively. 

\myparagraph{Evaluation metrics.} We report average top-1 accuracy across the four folds,  \emph{Manyshot} classes ($\geq100$ samples), \emph{Mediumshot} classes ($20\sim100$ samples), \emph{Fewshot} classes ($<20$ samples), and \emph{All} classes. Since the test set is balanced across all classes, the average accuracy and mean precision coincide. These four metrics are important for fine-grained evaluation since high accuracy on \emph{All} classes does not imply high accuracy on \emph{Fewshot} classes or \emph{Mediumshot} classes.

\myparagraph{Implementation Details.}
 For the \emph{Baseline} model, we take a Resnet-10 backbone for ImageNet-LT, following~\cite{liu2019large}. We initialise the model with Gaussian weights, use an initial learning rate of 0.2, and train for 100 epochs with a cosine learning rate schedule~\cite{loshchilov2016sgdr}. For Places-LT, we start with an ImageNet pre-trained Resnet-152 model, and finetune it with 0.01 learning rate for the first 30 epochs followed by 0.1 exponential decay in every 10 epochs. To train expert models, we initialise the feature extractor of each expert $\mathcal{E}_{\_}$ from the \emph{Baseline} model, and finetune it on its class-balanced subset. For $\mathcal{E}_\textit{Mediumshot}$ and $\mathcal{E}_\textit{Fewshot}$, we freeze the lower layers of the feature extractor and only learn the top few layers. The number of learnable layers is a hyperparameter that is fixed by measuring performance on the validation set. To train experts with the reject class, we fix the undersampling ratio for samples from the reject class by measuring performance on the validation set. Note that the classifier for each expert $\mathcal{E}_{\_}$ is smaller than the \emph{Baseline} model; it equals the number of classes in the expert's class-balanced subset, plus an additional reject class. 

\subsection{Oracle Performance}
To estimate the upper bound of our approach, we consider the performance with \emph{Oracle} access to expert selection information, that is, with apriori knowledge of the expert $\mathcal{E}_{\_}$ that contains the ground-truth class of a test sample in its class-balanced subset. The results are depicted in Table~\ref{tab:imagenet_oracle} and Table~\ref{tab:places_oracle}. The \emph{Oracle} outperforms the \emph{Baseline} by a significant margin on \emph{Mediumshot}, \emph{Fewshot} and \emph{All} accuracy. Moreover, it is significantly interesting to note that the Oracle accuracies on \emph{Mediumshot} and \emph{Fewshot} classes are on par with \emph{Manyshot} accuracy. This illustrates that performance drops on \emph{Mediumshot} and \emph{Fewshot} classes result from \emph{relative imbalance} rather than \emph{data scarcity}. Therefore, in principle, it is possible for a classification model to match \emph{Fewshot} and \emph{Mediumshot} accuracy with \emph{Manyshot} accuracy in modern large-scale datasets. It is also interesting to see that the \emph{Manyshot} accuracy does not improve much by using an \emph{Oracle}, suggesting that \emph{Manyshot} accuracy is already saturated in the \emph{Baseline} model.

\begin{table}[!]
\parbox{.48\linewidth}{
\small
\caption{Performance of Oracle vs Baseline on ImageNet-LT.}
\begin{tabular}{ l | c c c | c }
    \Xhline{4\arrayrulewidth}
    Method & Many & Medium & Few & All \\       \Xhline{4\arrayrulewidth}
    Baseline & \textbf{54.3} & 26.2 & 5.8 & 34.4 \\
    Experts (Oracle) & 54.2 & \textbf{43.3} & \textbf{45.7} & \textbf{47.9} \\  \Xhline{4\arrayrulewidth}
\end{tabular}
\label{tab:imagenet_oracle}
}
\hspace{2em}
\parbox{.45\linewidth}{
\small
\caption{Performance of Oracle vs Baseline on Places-LT.}
\begin{tabular}{ l | c c c | c }
    \Xhline{4\arrayrulewidth}
    Method & Many & Medium & Few & All \\       \Xhline{4\arrayrulewidth}
    Baseline & 45.4 & 25.6 & 9.0 & 29.5 \\
    Experts (Oracle) & \textbf{47.3} & \textbf{46.1} & \textbf{46.5} & \textbf{46.6} \\  \Xhline{4\arrayrulewidth}
\end{tabular}
\label{tab:places_oracle}
}
\end{table}

\subsection{Effect of joint calibration module}
We apply the methods outlined in Sec.~\ref{Fusing expert predictions} for fusing expert posteriors and compare their performance on ImageNet-LT and Places-LT. The results are depicted in Table~\ref{tab:imagenet_joint} and Table~\ref{tab:places_joint}. KL-div minimisation and Soft-voting yield the highest \emph{Fewshot} accuracy, however \emph{All} accuracy is much lower than the other methods. Expert selection and Stacking are better than KL-div minimisation and Soft-voting on \emph{Manyshot}, \emph{Mediumshot} and \emph{All} accuracy, but worse on \emph{Fewshot} accuracy. The Joint-calibration module obtains the best \emph{Manyshot}, \emph{Mediumshot} and \emph{All} accuracy, even though \emph{Fewshot} accuracy suffers.

\begin{table}[!]
\parbox{.45\linewidth}{
\caption{Effect of joint calibration module for ImageNet-LT.}
\begin{tabular}{ l | c c c | c }
    \Xhline{4\arrayrulewidth}
    Module & Many & Medium & Few & All \\       \Xhline{4\arrayrulewidth}
    KL-div min. & 25.3 & 20.5 & \textbf{39.1} & 21.9 \\ 
    Soft-Voting & 26.3 & 21.3 & 38.9 & 25.6 \\
    Expert Selection & 38.3 & 32.6 & 17.2 & 32.8 \\
    Stacking & 28.1 & 27.5 & 33.8 & 28.6 \\
    Joint Calibration & \textbf{43.2} & \textbf{34.3} & 18.9 & \textbf{35.7} \\  \Xhline{4\arrayrulewidth}
\end{tabular}
\label{tab:imagenet_joint}
}
\hspace{3em}
\parbox{.45\linewidth}{
\caption{Effect of joint calibration module for Places-LT. }
\begin{tabular}{ l | c c c | c }
     \Xhline{4\arrayrulewidth}
    Module & Many & Medium & Few & All \\       \Xhline{4\arrayrulewidth}
    KL-div min. & 30.2 & 31.7 & \textbf{28.9} & 30.4 \\
    Soft-Voting & 30.0 & 31.8 & 28.9 & 30.6 \\
    Expert Selection & 32.6 & 31.8 & 24.5 & 30.7 \\
    Stacking & 28.2 & 36.0 & 26.2 & 31.3 \\
    Joint Calibration & \textbf{37.2} & \textbf{35.3} & 26.3 & \textbf{34.2}  \\  \Xhline{4\arrayrulewidth}
\end{tabular}
\label{tab:places_joint}
}
\end{table}

\begin{figure*}[!t]
    \centering
    \begin{subfigure}[t]{0.49\textwidth}
        \includegraphics[width=\textwidth]{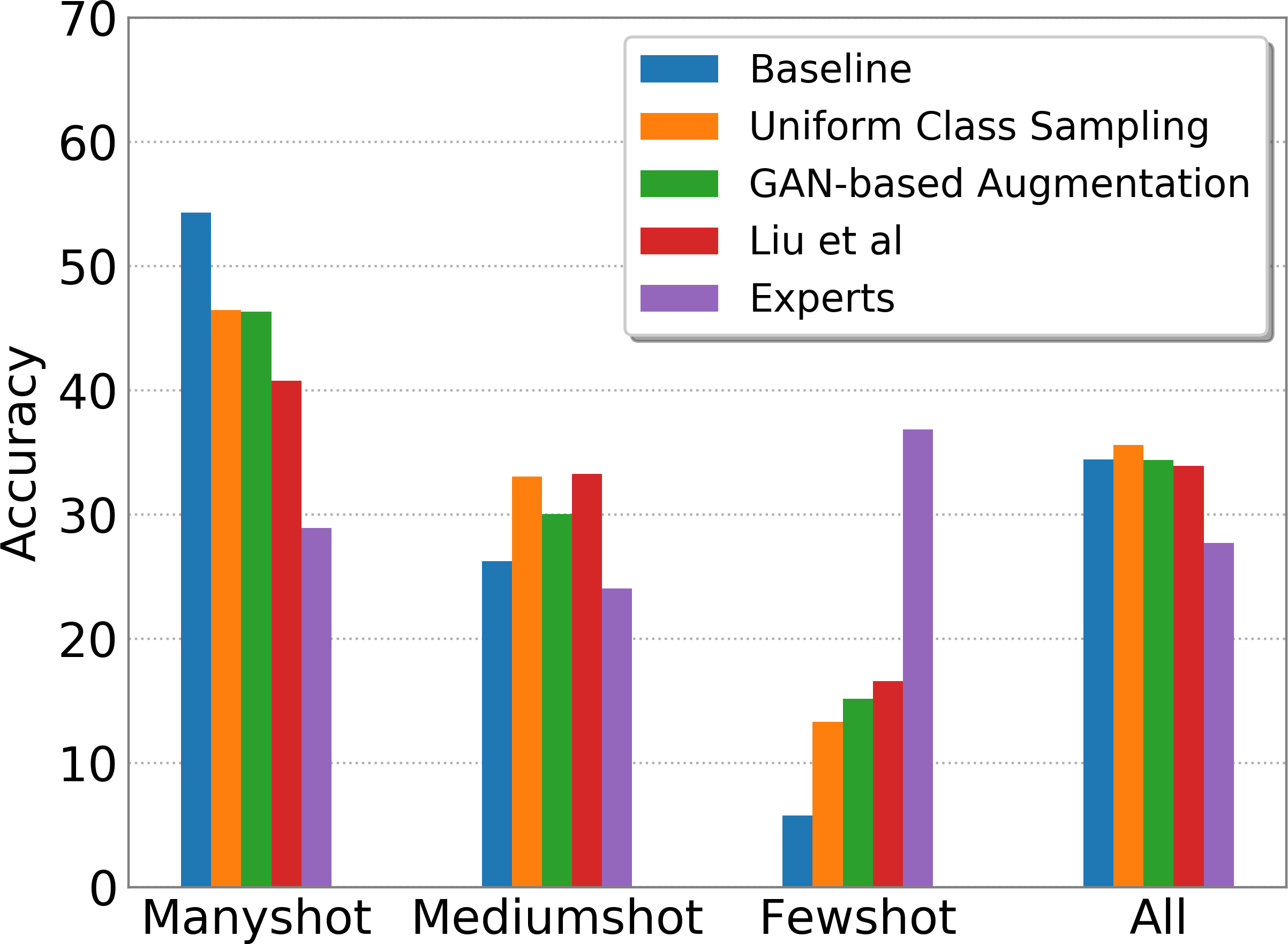}
        \caption{ImageNet-LT: Base Models}
        \label{fig:imagenet_base}
        \end{subfigure}
    \begin{subfigure}[t]{0.49\textwidth}
        \includegraphics[width=\textwidth]{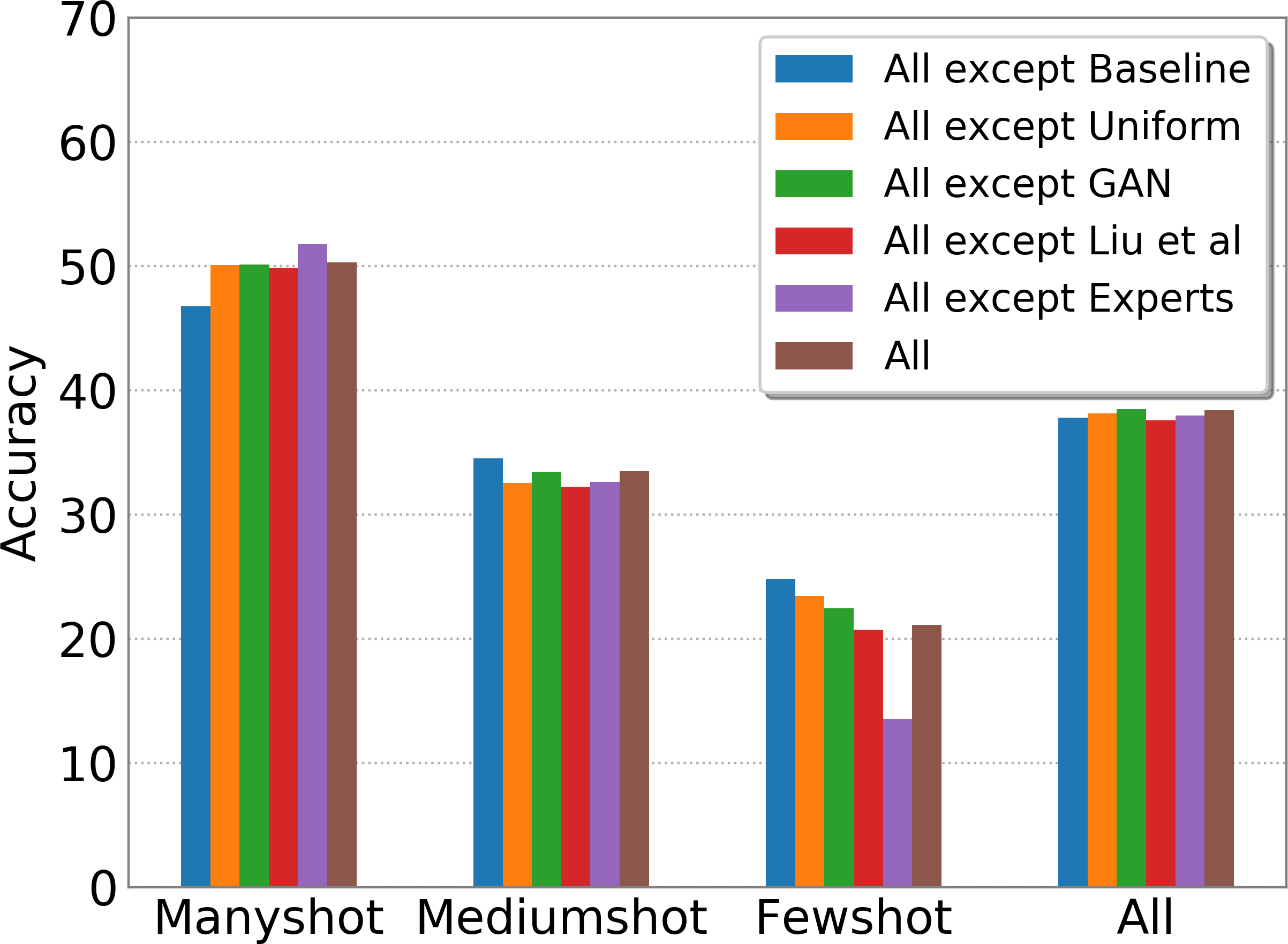} 
        \caption{ImageNet-LT: Take-One-Out Ensembles}
        \label{fig:imagenet_take1away}
        \end{subfigure}
    \begin{subfigure}[t]{0.49\textwidth}
        \includegraphics[width=\textwidth]{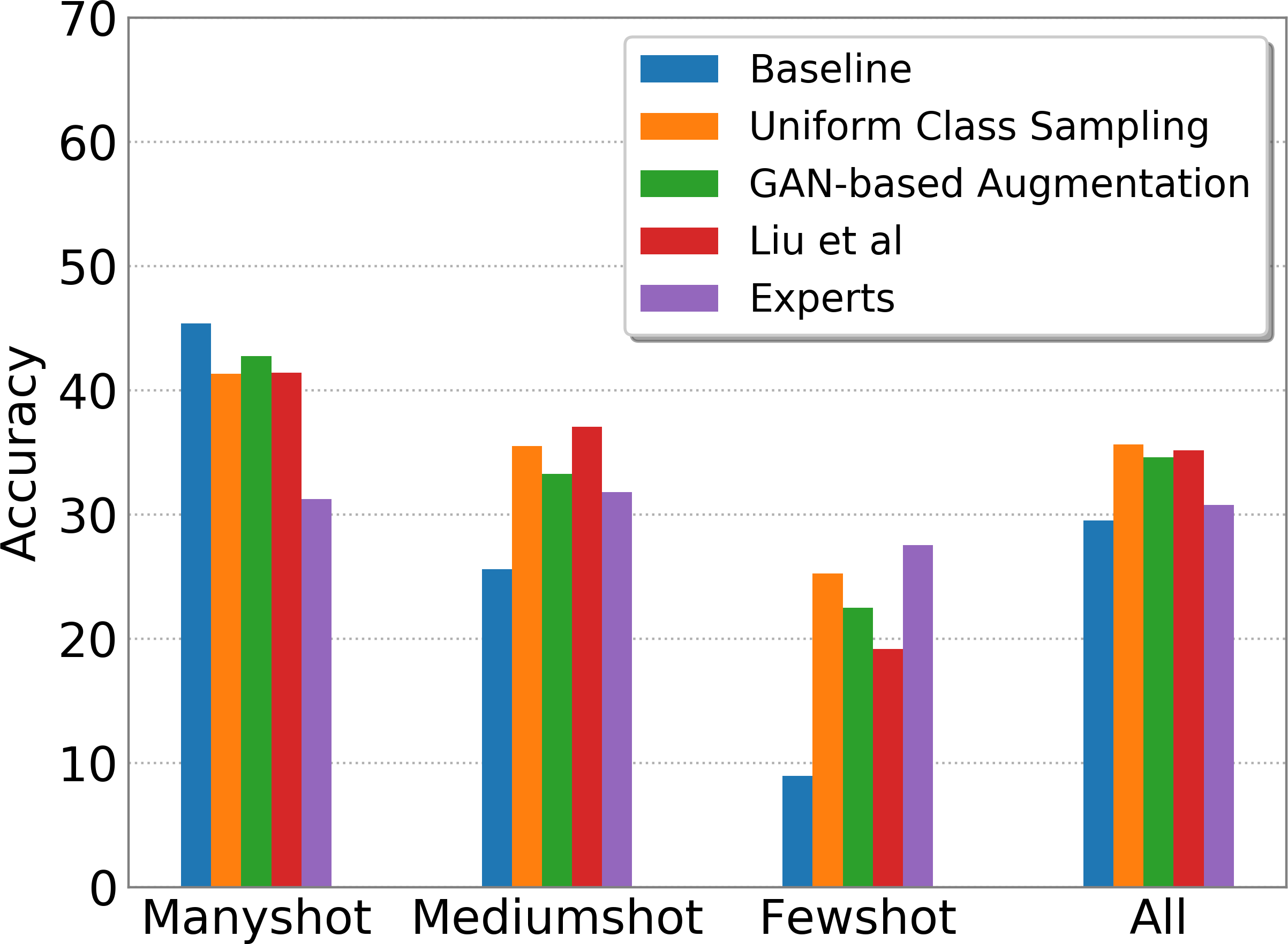}
        \caption{Places-LT: Base Models}
        \label{fig:places_base}
        \end{subfigure}
    \begin{subfigure}[t]{0.49\textwidth}
        \includegraphics[width=\textwidth]{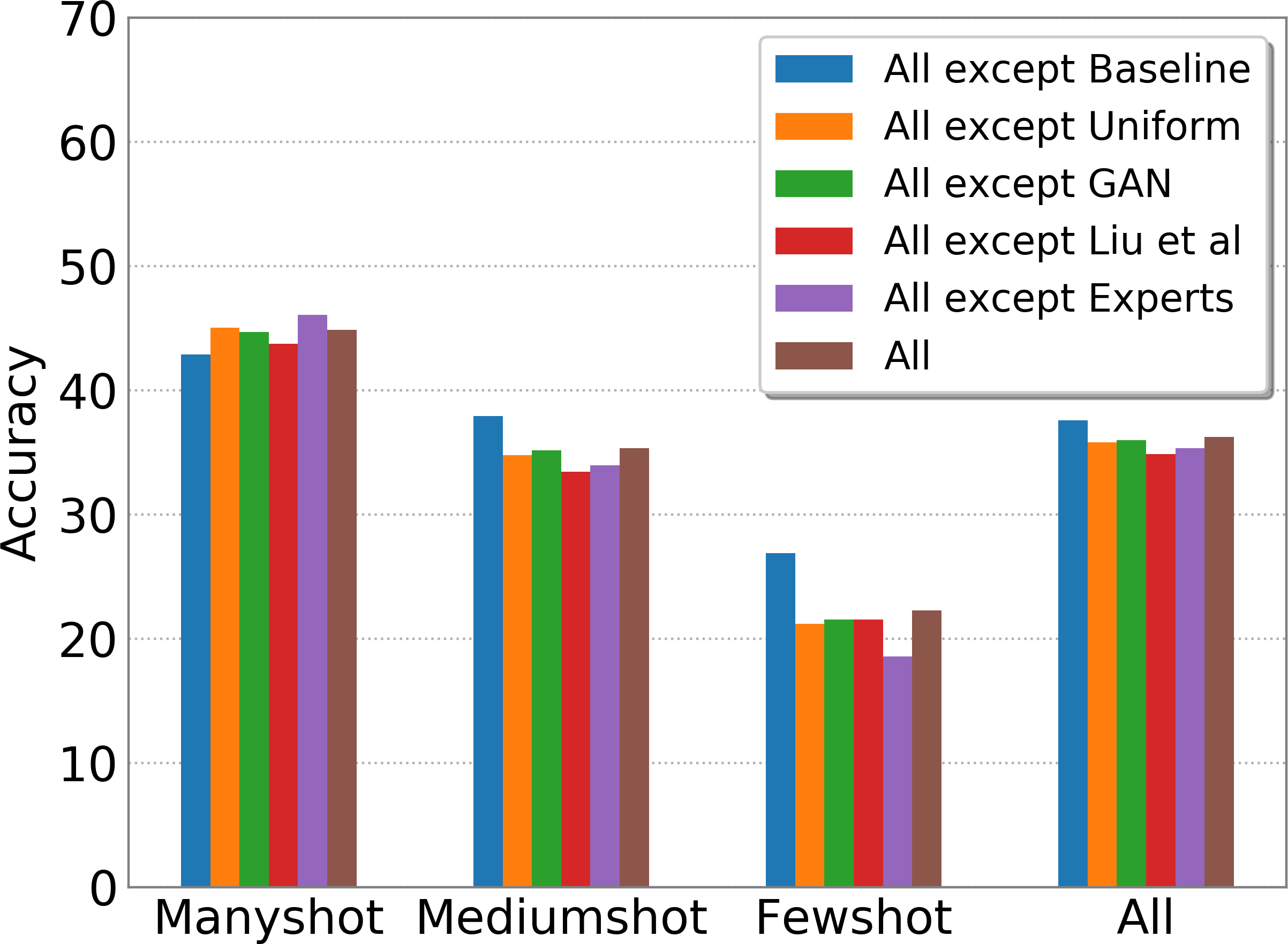} 
        \caption{Places-LT: Take-One-Out Ensembles}
        \label{fig:places_take1away}
        \end{subfigure}
   \caption{From L-R: Performance of - Base Models, and Take-One-Out ensembles. All results are evaluated on the testing set. Top and bottom rows correspond to ImageNet-LT and Places-LT respectively. Best viewed in color with zoom.} 
    \label{fig:ensemblecombination_ablation}
    \vspace{-4mm}
\end{figure*}

\subsection{Diverse ensembles with experts}
In this section, we extend our ensemble using existing long-tailed recognition solutions and analyse the performance of various combinations of models in the ensemble. We experiment with the following models: (i) The \emph{Baseline} model, (ii) The three expert models, $\mathcal{E}_\textit{Manyshot}$, $\mathcal{E}_\textit{Mediumshot}$ and $\mathcal{E}_\textit{Fewshot}$ fused using Soft-voting, collectively referred to as \emph{Experts}, (iii) Classifier finetuning with uniform class sampling, wherein we freeze the feature extractor of the \emph{Baseline} model and finetune the classifier with uniform class sampling. This is referred to as \emph{Uniform class sampling} or \emph{Uniform}, (iv) Data augmentation for \emph{Mediumshot} and \emph{Fewshot} classes using a conditional generative model from class embeddings to feature embeddings, denoted as \emph{GAN based augmentation} or simply \emph{GAN}, (v) Knowledge transfer from \emph{Manyshot} to \emph{Fewshot} classes using a learned convex combination of class prototypes from \cite{liu2019large}, denoted as \emph{Liu et. al.}. The performances of these base models are depicted in Fig.~\ref{fig:imagenet_base} and Fig.~\ref{fig:places_base}. Notice how the performance of the \emph{Baseline} model degrades from \emph{Manyshot} to \emph{Mediumshot} to \emph{Fewshot} accuracy. The \emph{Expert} models give the highest accuracy on the \emph{Fewshot} classes, but are worse on \emph{Manyshot} accuracy. 

\par We combine all these models into a single ensemble, take one model out and see the effect on the performance. To keep the analysis simple, we use Soft-voting for fusing posteriors from all the models, since it doesn't involve learning additional parameters. This ablation is depicted in Fig.~\ref{fig:imagenet_take1away} and Fig.~\ref{fig:places_take1away}. As expected, the diverse ensembles give higher \emph{All} accuracy than the base models. Taking \emph{Experts} out causes performance drop on \emph{Mediumshot}, \emph{Fewshot} and \emph{All} accuracy, and increase in accuracy on \emph{Manyshot} classes. This suggests that the \emph{Experts} are important in the ensemble for high \emph{Mediumshot} and \emph{Fewshot} accuracy. On the other hand, taking the \emph{Baseline} model out of the ensemble causes an increase in \emph{Fewshot} accuracy while \emph{Manyshot} accuracy drops. The ablation also reveals the inherent trade-off between \emph{Manyshot} and \emph{Fewshot} accuracy; an appropriate combination of models can tilt accuracy in favor of \emph{Manyshot} or \emph{Fewshot} classes. 

\begin{table}[!]
\parbox{.45\linewidth}{
\caption{Results on ImageNet-LT, using backbone Resnet-10. *Results obtained from the author's code. \ddag Results taken directly from~\cite{liu2019large}.}
\begin{tabular}{ l | c c c | c }
    \Xhline{4\arrayrulewidth}
  Methods & Many & Medium & Few & All \\      \Xhline{4\arrayrulewidth}
    Lifted Loss\ddag~\cite{oh2016deep} & 35.8 & 30.4 & 17.9 & 30.8 \\
    Focal Loss\ddag~\cite{lin2017focal} & 36.4 & 29.9 & 16 & 30.5 \\
    Range Loss\ddag~\cite{zhang2017range} & 35.8 & 30.3 & 17.6 & 30.7 \\
    FSLwF\ddag~\cite{gidaris2018dynamic} & 40.9 & 22.1 & 15 & 28.4 \\
    Liu~et~al.\ddag~\cite{liu2019large} & 43.2 & 35.1 & 18.5 & 35.6 \\ \hline
    Baseline & \textbf{54.3} & 26.2 & 5.7 & 34.4 \\
    Uniform & 46.5 & 33.0 & 13.3 & 35.6 \\
    GAN & 46.4 & 30.0 & 15.2 & 34.4 \\
    Liu~et~al.*~\cite{liu2019large} & 40.8 & 33.3 & 16.6 & 33.9 \\           
    \textbf{Ours (\emph{Experts})} & 43.2 & 34.3 & 18.9 & 35.7 \\
  \textbf{Ours (\emph{All})} & 48.2 & \textbf{37.0} & \textbf{21.5} & \textbf{39.2} \\      \Xhline{4\arrayrulewidth}
\end{tabular}
\label{tab:imagenet_results}
}
\hspace{3em}
\parbox{.45\linewidth}{
\caption{Results on Places-LT, using backbone Resnet-152. *Results obtained from the author's code. \ddag Results taken directly from~\cite{liu2019large}.}
\begin{tabular}{ l | c c c | c }
    \Xhline{4\arrayrulewidth}
  Methods & Many & Medium & Fews & All \\      \Xhline{4\arrayrulewidth}
  Lifted Loss\ddag~\cite{oh2016deep}  & 41.1 & 35.4 & 24.0 & 35.2 \\      
  Focal Loss\ddag~\cite{lin2017focal}  & 41.1 & 34.8 & 22.4 & 34.6 \\ 
  Range Loss\ddag~\cite{zhang2017range}  & 41.1 & 35.4 & 23.2 & 35.1 \\ 
  FSLwF\ddag~\cite{gidaris2018dynamic}  & 43.9 & 29.9 & \textbf{29.5} & 34.9 \\       
  Liu~et~al.\ddag~\cite{liu2019large} & 44.7 & 37.0 & 25.3 & 35.9 \\     \hline
  Baseline  & \textbf{45.4} & 25.6 &  9.0 & 29.5  \\      
  Uniform & 41.3 & 35.5 & 25.2 & 35.6 \\
  GAN & 42.7 & 33.3 & 22.5 & 34.6 \\
  Liu~et~al.*~\cite{liu2019large} & 41.4 & 37.1 & 19.2 & 35.2 \\             
  \textbf{Ours (\emph{Experts})} & 37.2 & 35.3 & 26.3 & 34.2 \\
  \textbf{Ours (\emph{All})} & 43.6 & \textbf{39.9} & 27.7 & \textbf{38.9} \\      \Xhline{4\arrayrulewidth}
\end{tabular}
\label{tab:places_results}
}
\end{table}

\subsection{Comparison to the state-of-the-art}
We now compare our ensemble of class-balanced experts and the diverse ensemble described in the previous section to the state-of-the-art on the test set of ImageNet-LT and Places-LT. All ensemble combinations use the joint calibration module to fuse model posteriors as it gives us the highest average accuracy. The results are depicted in Table~\ref{tab:imagenet_results} and Table~\ref{tab:places_results}. We observe that Ours (Experts) gives us close to state-of-the-art results, and Ours (All) establishes a new state-of-the-art on both the benchmark datasets. This validates our hypothesis that an ensemble of class-balanced expert models is a simple and effective strategy for dealing with long-tailed datasets.

\subsection{Discussion}

\begin{figure*}[!t]
        \vspace{-1em}
        \begin{subfigure}{0.32\linewidth}
        \includegraphics[width=\linewidth]{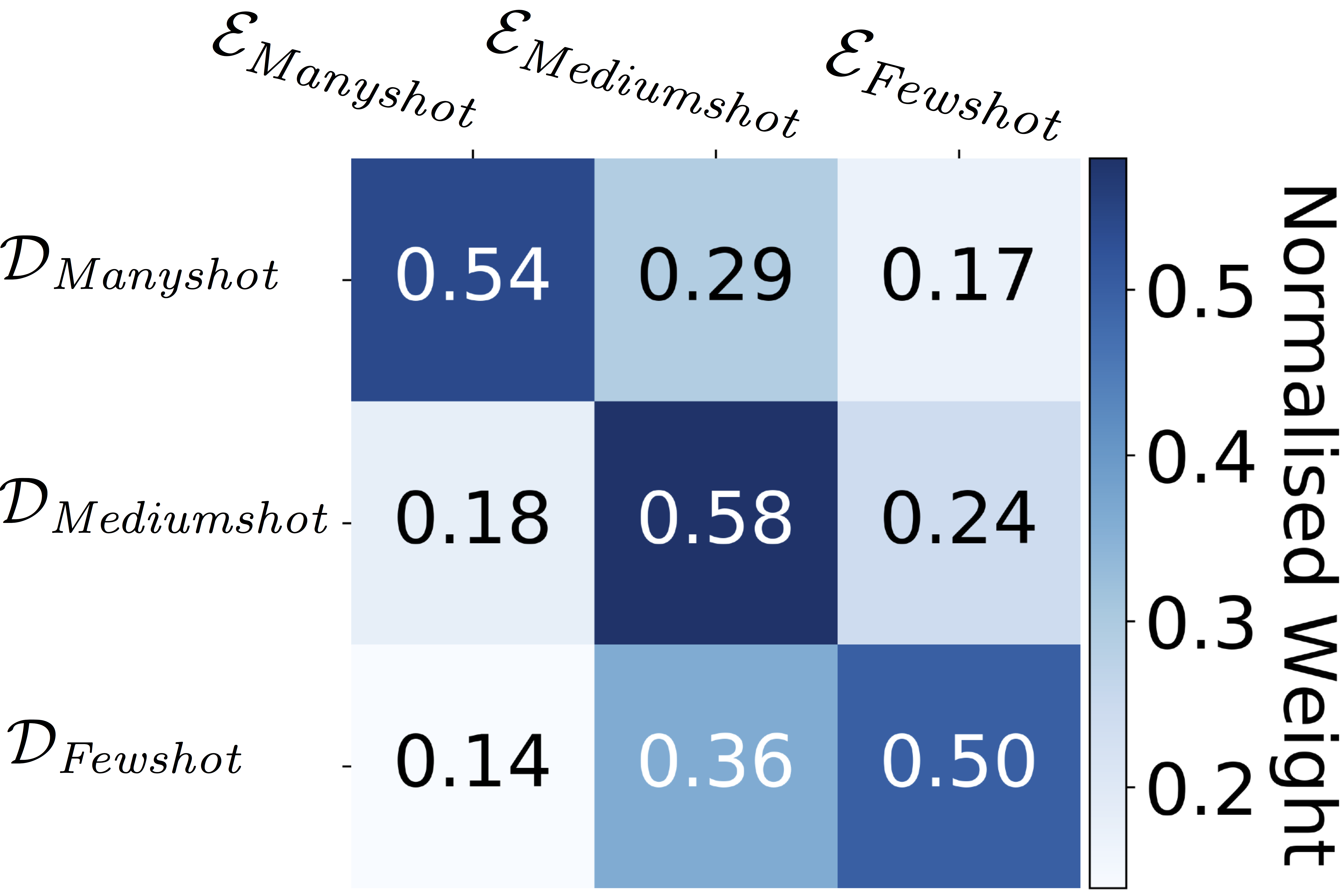}
        \caption{}
        \label{fig:expert collision}
        \end{subfigure}
    \begin{subfigure}{0.32\linewidth}
        \includegraphics[width=\textwidth]{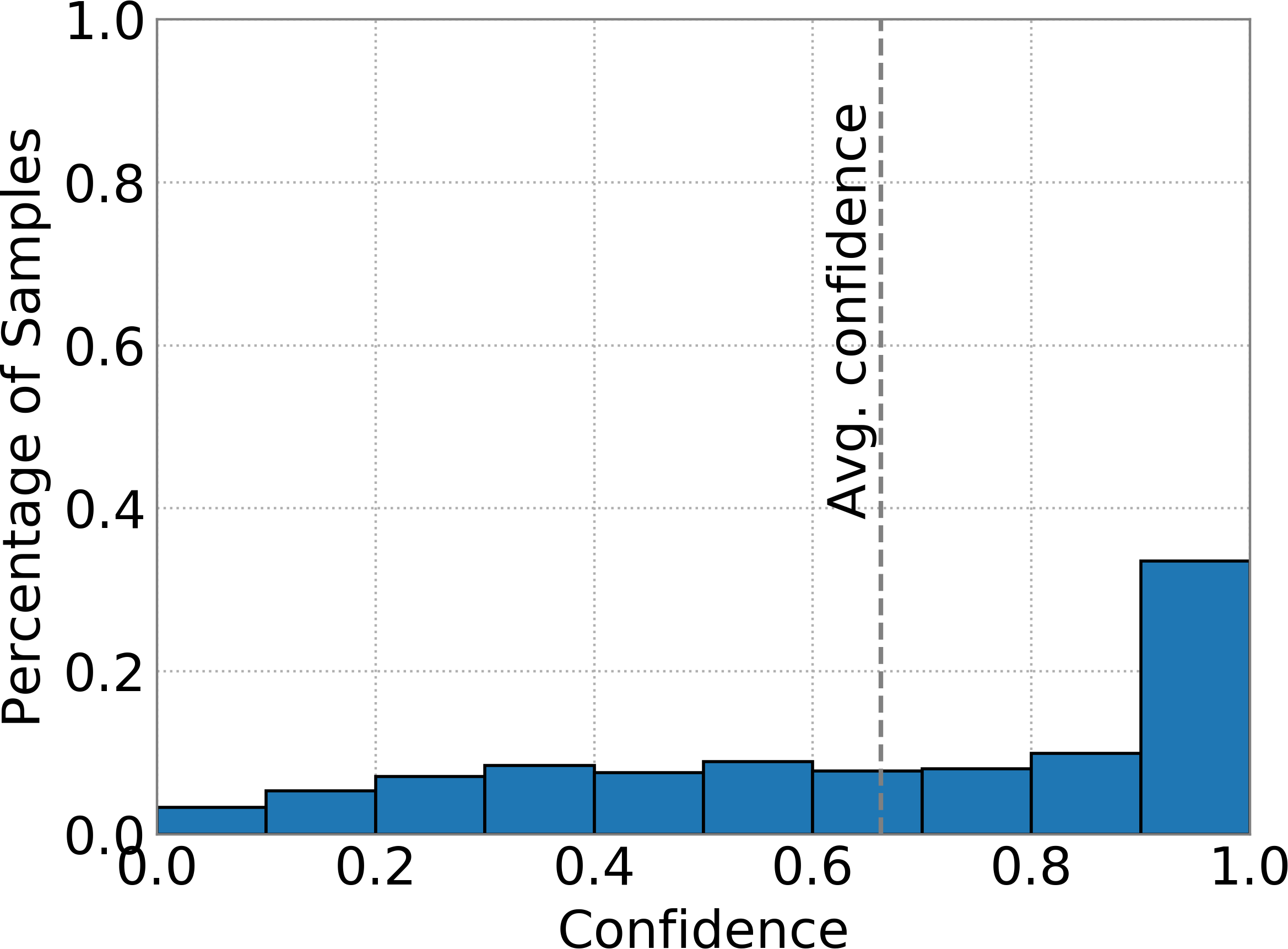}
        \caption{}
        \label{fig:manyshotexperthist}
        \end{subfigure}
    \begin{subfigure}{0.32\linewidth}
        \includegraphics[width=\textwidth]{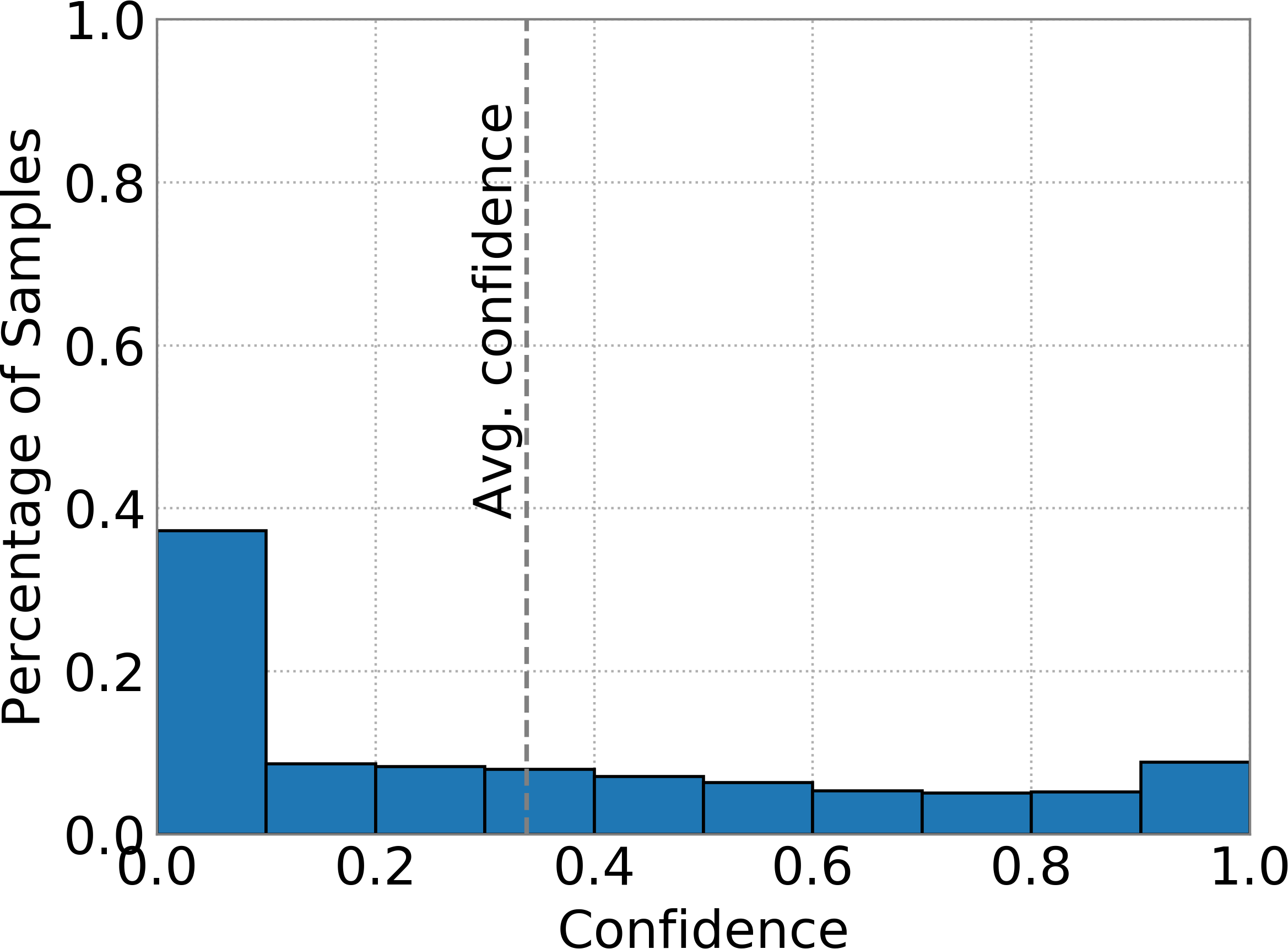}
        \caption{}
        \label{fig:fewshotexperthist}
        \end{subfigure}
     \begin{subfigure}{0.32\linewidth}
        \includegraphics[width=\linewidth]{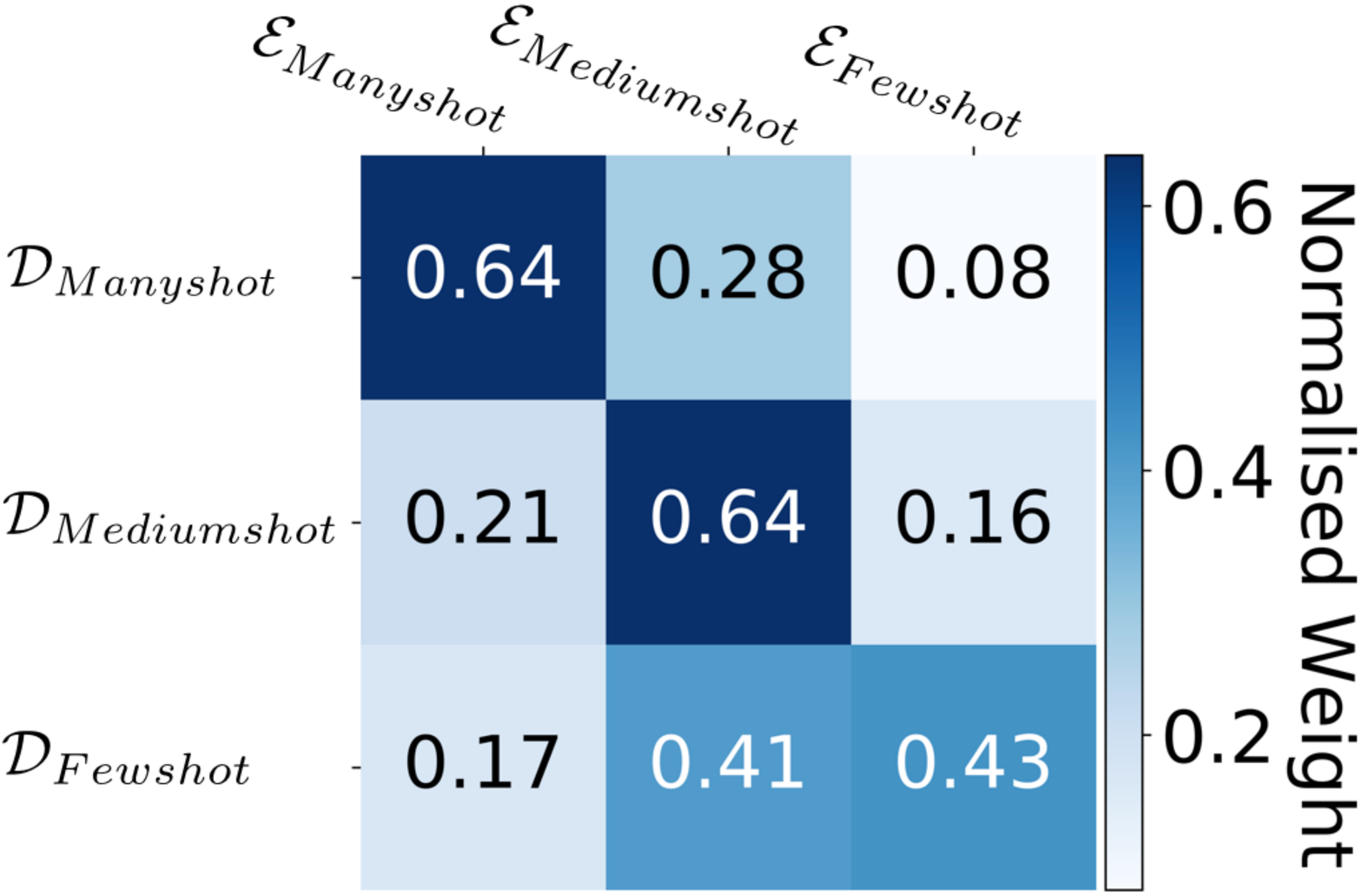}
        \caption{}
        \label{fig:expert collision_jc}
        \end{subfigure}
        \hspace{0.2em}
    \begin{subfigure}{0.32\linewidth}
        \includegraphics[width=\textwidth]{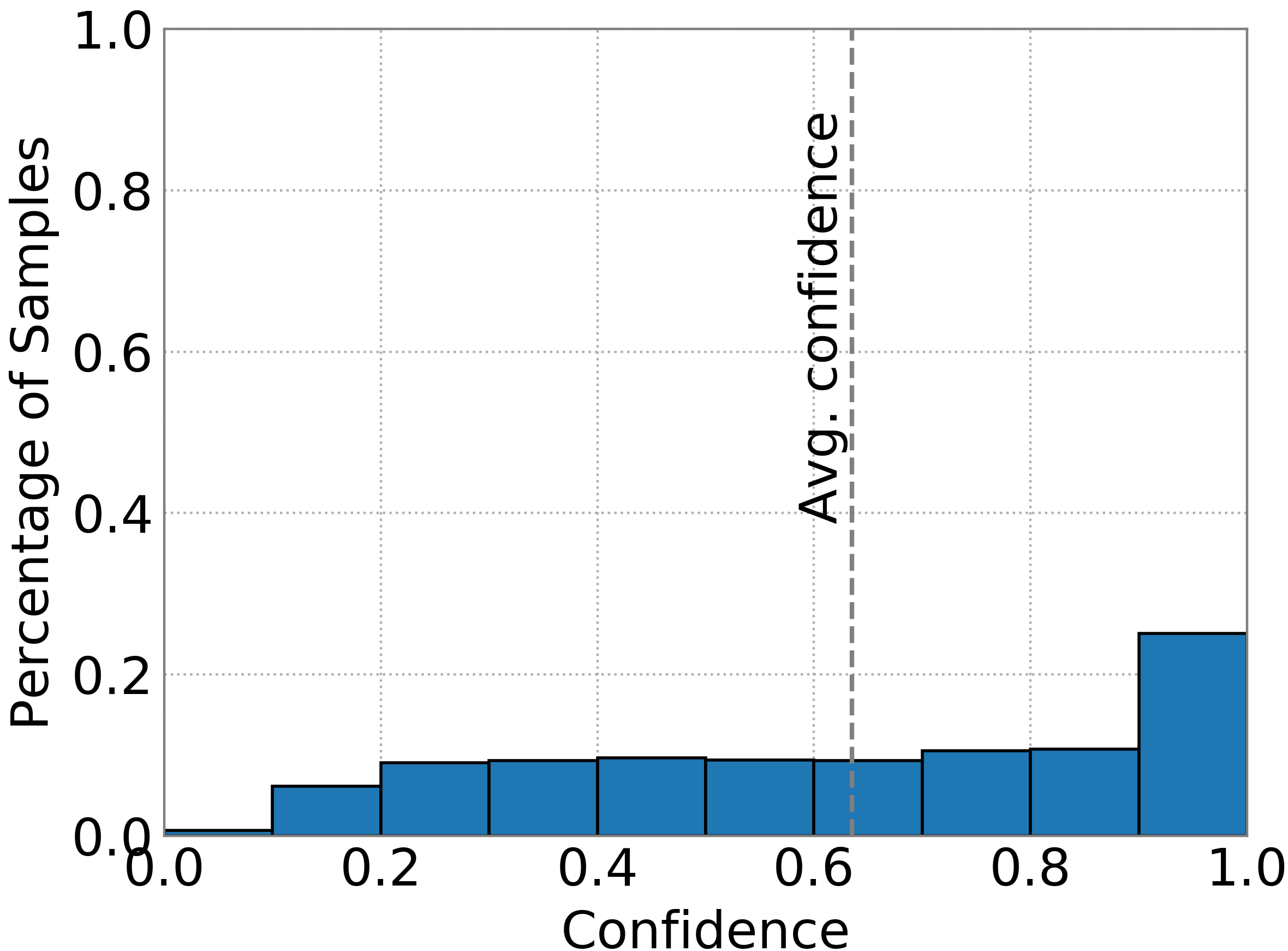}
        \caption{}
        \label{fig:manyshotexperthist_jc}
        \end{subfigure}
        \hspace{0.2em}
    \begin{subfigure}{0.32\linewidth}
        \includegraphics[width=\textwidth]{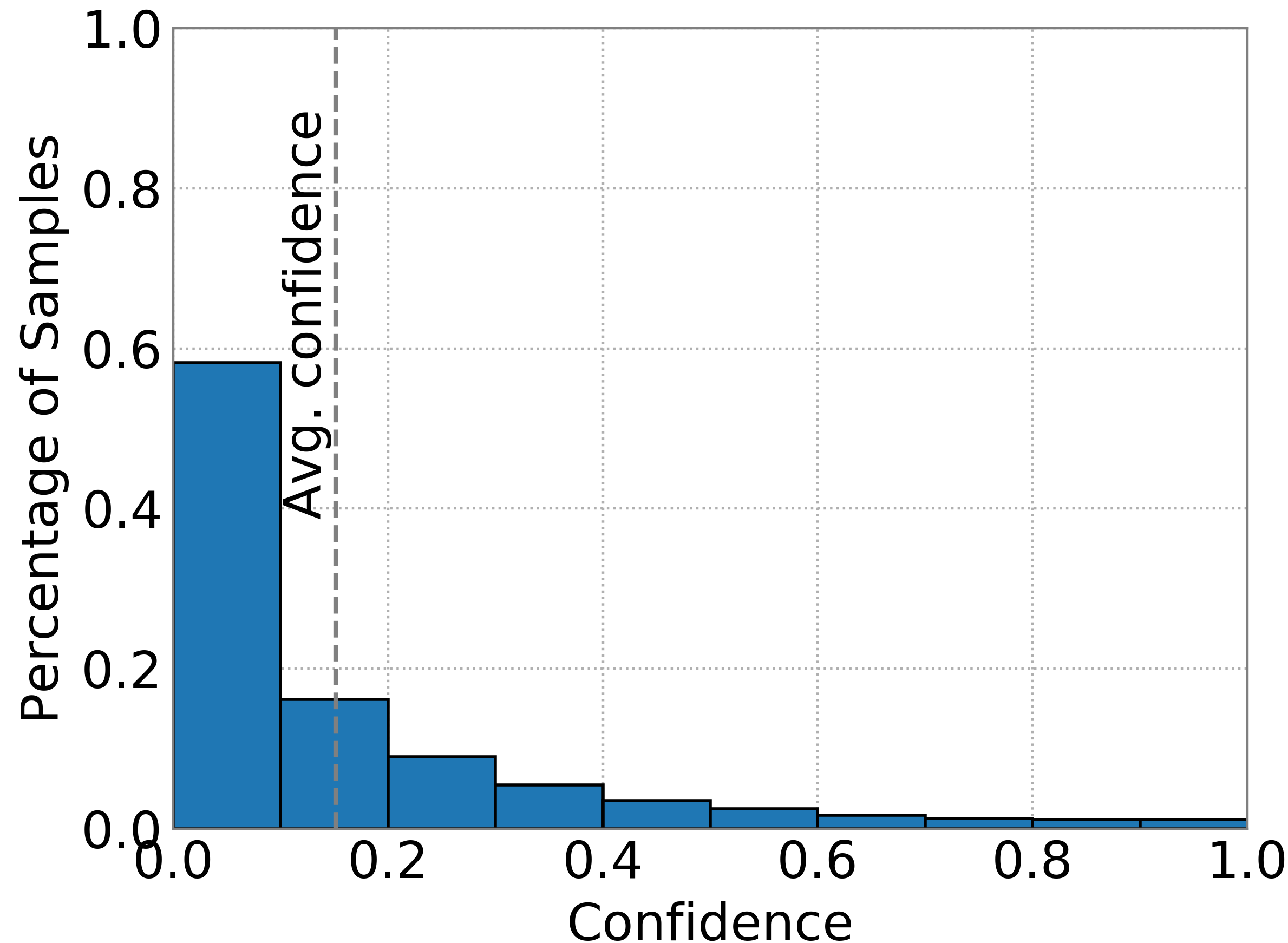}
        \caption{}
        \label{fig:fewshotexperthist_jc}
        \end{subfigure}
    \caption{Top (bottom): Before (after) joint calibration. L-R: Expert confusion matrix, confidence histograms of $\mathcal{E}_\textit{Manyshot}$ for samples it correctly classifies in $\mathcal{E}_\textit{Manyshot}$, and $\mathcal{E}_\textit{Fewshot}$ for the same samples. All results on Places-LT. Joint calibration aligns experts' confidences and decreases \emph{expert collision}.}
    \label{fig:discussion}
\end{figure*}

There is significant difference between the results depicted in Table~\ref{tab:imagenet_oracle} and Table~\ref{tab:places_oracle}, and Table~\ref{tab:imagenet_results} and Table~\ref{tab:places_results}. This shows that the various strategies used for fusing expert posteriors are sub-optimal. To analyse the underlying cause, we take our ensemble of class-balanced experts and plot a confusion matrix, each entry showing the percentage of samples from dataset $\mathcal{D}_{\_}$ that are classified by expert model $\mathcal{E}_{\_}$. For the preliminary analysis we use Soft-voting for fusing expert posteriors. Fig.~\ref{fig:expert collision} shows the result for Places-LT. The plot shows there is significant confusion amongst experts; experts aren't selected optimally for classes to which a test sample belongs. We term this phenomenon as \emph{Expert collision}.

\par We further consider each expert's confidence in its predictions. We take the confidence or the maximum softmax probability (MSP) from the expert posteriors and plot confidence histograms. We do this for $\mathcal{E}_\textit{Manyshot}$ on its class-balanced subset $\mathcal{D}_\textit{Manyshot}$, for samples from the test set it correctly classifies, and for $\mathcal{E}_\textit{Fewshot}$ on the same test samples from $\mathcal{D}_\textit{Manyshot}$. This is depicted in Fig.~\ref{fig:manyshotexperthist} and Fig.~\ref{fig:fewshotexperthist}. The plots show that $\mathcal{E}_\textit{Manyshot}$ has high confidence predictions while $\mathcal{E}_\textit{Fewshot}$ has low confidence predictions on these samples. However, to avoid \emph{Expert collision} both the confidence histograms should have a reasonable margin in between and not overlap. Fig.~\ref{fig:expert collision_jc} and Fig.~\ref{fig:manyshotexperthist_jc},~\ref{fig:fewshotexperthist_jc} respectively show the confusion matrix and confidence histograms after joint calibration. It's essential to align confidences of the three experts correctly, and this is precisely what \emph{joint calibration} does by learning scale and shift parameters for each class. 
\section{Conclusion}
This article presented an ensemble of class-balanced experts framework for long-tailed recognition. Our effective and modular strategy explicitly tackles \emph{relative imbalance} without resorting to complex models or sophisticated loss objectives. We decompose the imbalanced classification problem into balanced classification problems that are more tractable, and train separate expert models for \emph{Manyshot}, \emph{Mediumshot} and \emph{Fewshot} subsets of the data with a reject class for samples lying outside an expert's class-balanced subset. We scale and shift experts' partial posteriors to jointly calibrate experts' predictions, and our ensemble of class-balanced experts reaches close to state-of-the-art performance on two long-tailed benchmarks. We also extend our ensemble with diverse existing solutions for long-tailed recognition and establish a new state-of-the-art on the two benchmark datasets. Moreover, our experiments with an Oracle upper bound reveal that performance drops on \emph{Mediumshot} accuracy and \emph{Fewshot} accuracy are caused by \emph{relative imbalance} and not \emph{data scarcity} for rare classes. Therefore, it is possible to bring \emph{Mediumshot} and \emph{Fewshot} accuracy on par with \emph{Manyshot} accuracy by remedying \emph{relative imbalance} in modern large-scale datasets, which motivates further research in this direction.

\bibliographystyle{splncs04}
\bibliography{egbib}

\end{document}